%% file: main.tex
\documentclass[letterpaper]{article} 
\usepackage{aaai25}  
\usepackage{times}  
\usepackage{helvet}  
\usepackage{courier}  
\usepackage[hyphens]{url}  
\usepackage{graphicx} 
\usepackage{natbib}  
\usepackage{caption} 
\usepackage{algorithm}
\usepackage{algorithmic}

\usepackage{newfloat}
\usepackage{listings}
\DeclareCaptionStyle{ruled}{labelfont=normalfont,labelsep=colon,strut=off} 
\floatstyle{ruled}
\newfloat{listing}{tb}{lst}{}
\floatname{listing}{Listing}
\usepackage{epsfig}
\usepackage{amsmath}
\usepackage{amssymb}

\usepackage{flexisym}
\usepackage{multirow}
\usepackage{subcaption}

\usepackage{tikz}
\usepackage{comment}

\usepackage{dsfont}
\usepackage{color, colortbl}

\usepackage{bbold}
\usepackage{mathtools}
\usepackage{kotex}

\usepackage[accsupp]{axessibility}  

\DeclareMathOperator*{\argmin}{arg\,min}

\makeatletter
\def\@fnsymbol#1{\ensuremath{\ifcase#1\or \dagger\or \ddagger\or
\mathsection\or \mathparagraph\or \|\or **\or \dagger\dagger
\or \ddagger\ddagger \else\@ctrerr\fi}}
\makeatother

\title{Prediction-Feedback DETR for Temporal Action Detection}
\author{
    Jihwan Kim~~ Miso Lee~~ Cheol-Ho Cho~~ Jihyun Lee~~ Jae-Pil Heo\thanks{Corresponding author}
}
\affiliations{
    {\Large
    Sungkyunkwan University}\\
    \vspace{3.0pt}
    {\tt\small \{damien,\;dlalth557,\;gersys,\;ibluee01,\;jaepilheo\}@skku.edu}
}

\usepackage{bibentry}

\begin{document}

\maketitle

\input{sections/0_abstract}

%

\input{sections/1_introduction}
\input{sections/2_related_work}
\input{sections/3_our_approach}
\input{sections/4_experiments}
\input{sections/5_conclusion}

\section*{Acknowledgements}
This work was supported in part by MSIT/IITP (No. 2022-0-00680, 2020-0-01821, 2019-0-00421, RS-2024-00459618, RS-2024- 00360227, RS-2024-00437102, RS-2024-00437633), and MSIT/NRF (No. RS-2024-00357729).

\bibliography{aaai25}

\newpage

\input{sections/7_supplementary}

\end{document}

%% file: sections/0_abstract.tex
\begin{abstract}
Temporal Action Detection (TAD) is fundamental yet challenging for real-world video applications.
Leveraging the unique benefits of transformers, various DETR-based approaches have been adopted in TAD.
However, it has recently been identified that the attention collapse in self-attention causes the performance degradation of DETR for TAD.
Building upon previous research, this paper newly addresses the attention collapse problem in cross-attention within DETR-based TAD methods.
Moreover, our findings reveal that cross-attention exhibits patterns distinct from predictions, indicating a short-cut phenomenon.
To resolve this, we propose a new framework, Prediction-Feedback DETR (Pred-DETR), which utilizes predictions to restore the collapse and align the cross- and self-attention with predictions.
Specifically, we devise novel prediction-feedback objectives using guidance from the relations of the predictions.
As a result, Pred-DETR significantly alleviates the collapse and achieves state-of-the-art performance among DETR-based methods on various challenging benchmarks including THUMOS14, ActivityNet-v1.3, HACS, and FineAction.
\end{abstract}

%% file: sections/1_introduction.tex
\section{Introduction}
With the advancement of society, the use of video media has become increasingly widespread. 
As a result, the demand for efficient methods to search for desired segments within untrimmed videos has grown significantly. 
One fundamental task, Temporal Action Detection (TAD) aims to identify specific actions within a video and determine their temporal boundaries. 
TAD has primarily advanced through two-stage approaches. 
However, recent research has increasingly focused on end-to-end DETR-based methods.

DETR~\cite{carion2020detr} is a framework initially proposed and developed in the literature of object detection, introducing the first end-to-end detection framework using set prediction. 
The DETR method has also been extended to the video domain and applied to TAD~\cite{tan2021relaxed, liu2022tadtr, shi2022react}. 
In TAD, each query is used to predict an action within the video along with its corresponding time interval. 
To achieve this, bipartite matching is employed to align each query with the ground truth actions and their temporal intervals within the untrimmed video. 
This approach has the distinct advantage of eliminating traditional heuristics like Non-Maximum Suppression (NMS).

\begin{figure}[t]
\centering
\includegraphics[width=8.35cm]{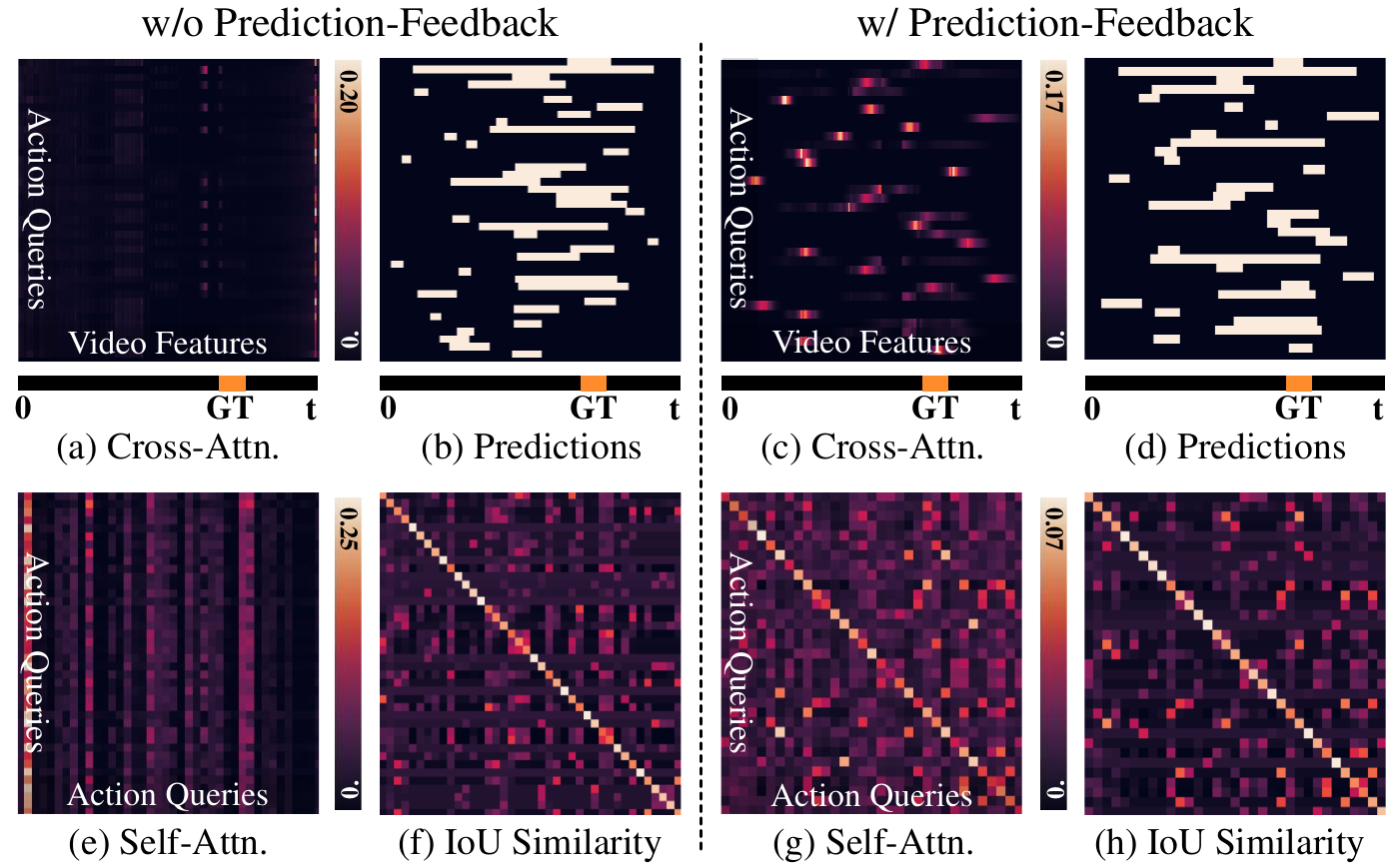}
\caption{\textbf{Attention collapse problem. }
The figure depicts the cross- ((a), (c)) and self-attention maps ((e), (g)) of the decoder as well as the predictions ((b), (d)) and their normalized IoU similarity map ((f), (h)).
DETR for TAD with standard attention severely suffers from the attention collapse in its cross-attention and self-attention ((a), (e)).
The collapsed attention focuses on a few encoder features (a) or decoder queries (e) regardless of the DETR predictions ((b), (f)).
}
\label{fig:introduction}
\end{figure}

Although DETR with standard attention (shortly, original-DETR) has advanced compatibly with Deformable-DETR~\cite{xizhou2021deformable_detr} in object detection, original-DETR in TAD even with recent architectures like DAB-DETR~\cite{shilong2022dab_detr} shows a way worse performance.
Recently, the root of this issue is identified as the attention collapse problem in self-attention (SA) by Self-DETR~\cite{kim2023self} as depicted in (e) of Fig.~\ref{fig:introduction}, where all decoder queries focus on a few queries.
The attention collapse is the phenomenon of skipping the attention module to prevent from degeneration of the model~\cite{dong2021rank_collapse} towards a rank-1 matrix.
Self-DETR utilizes the cross-attention (CA) map to recover the collapsed SA.

However, their solution depends on the soundness of the CA, otherwise it could be sub-optimal.
We discover that it is not sound; rather collapsed as depicted in Fig.~\ref{fig:introduction}.
The figure shows that the decoder queries in CA attend to a few encoder features ((a) of the figure), exhibiting the same pattern over almost all queries.
It is a particularly critical issue because CA is crucial for the task as it bridges the queries and the video features.
This leads us to resolve the collapse of CA and develop another approach for self-feedback.

Fig.~\ref{fig:introduction} also illustrates the localization predictions in (b), and their corresponding Interaction-over-Union (IoU) map as the self-relation of the queries in (f).
In the figure, the attention maps demonstrate clearly different patterns regardless of their predictions and self-relation.
Typically, we interpret that the attention maps represent where the model focuses, thus implying why it produces those results.
Hence, this phenomenon is analogous to a shortcut, where the model relies on simpler cues rather than learning meaningful representations.
Despite this collapsed attention, the model still generates diverse and plausible results, even though all queries focus on the same background regions, as seen in (a) of the CA.
This occurs because bipartite matching in the objective of DETR enforces varied predictions by penalizing duplicate results.
Based on this observation, we suggest that attention maps be aligned with their corresponding predictions.
By using the predictions, rather than the collapsed CA, as a guide for attention, we aim to generalize the model and address the issue of attention collapse.

To this end, we propose a new framework, Prediction-Feedback DETR (Pred-DETR), to tackle the collapse of the entire attention mechanisms in DETR.
Our approach begins by expressing the relation of the decoder queries as the IoU similarity map of the DETR predictions with their time intervals.
We also reformulate the CA map into the self-relation of the decoder queries.
Next, we introduce an auxiliary objective that aligns the self-relation from the CA and SA maps with the IoU similarity map derived from the predictions.
Additionally, we leverage encoder predictions from the recent DETR mechanism to guide the encoder SA and decoder CA.
Through extensive experiments with various challenging benchmarks including THUMOS14, ActivityNet-v1.3, HACS, and FineAction, 
we demonstrate that the proposed methods remarkably reduce the degree of the attention collapse problem.
Furthermore, the activated attention leads to substantial performance improvements, achieving a new state-of-the-art among DETR-based methods.

To sum up, our main contributions are as follows:
\begin{itemize}
	\item We identify the attention collapse problem in cross-attention of DETR for TAD.
    Especially, we found that the cross-attention exhibits clearly different patterns from the predictions, which implies a short-cut phenomenon due to the collapse.
	\item We propose a novel framework, Prediction-Feedback DETR (Pred-DETR), which utilizes predictions for relieving the attention collapse.
    We give an auxiliary objective for the collapsed attention modules to be aligned with the IoU relation of the predictions.
	\item Our extensive experiments demonstrate that Pred-DETR remarkably reduces the degree of the attention collapse by maintaining high diversity of attention. Moreover, we validate that our model achieves a new state-of-the-art performance over the DETR-based models on THUMOS14, ActivityNet-v1.3, HACS, and FineAction.
\end{itemize}

%% file: sections/2_related_work.tex
\section{Related Work}
\subsection{Temporal Action Detection}
Temporal action detection (TAD) task focuses on identifying time intervals of action and classifying the instance within untrimmed videos.
Over the past decade, significant advancements in TAD have been achieved  by foundational methods~\cite{yeung2016frame-glimpses, shou2016scnn, buch2017sst}.
Inspired by the success of two-stage mechanisms in object detection, many TAD methods have adopted a multi-stage framework~\cite{gao2017turn, zhao2017ssn, xu2017rc3d, kim2019coarsefine}.

As the subsequent work, point-wise learning has been widely adopted to generate more flexible proposals without pre-defined time windows.
SSN~\cite{zhao2017ssn} and TCN~\cite{dai2017tcn} introduced extended temporal context around the generated proposals to enhance ranking performance.
BSN~\cite{lin2018bsn} and BMN~\cite{lin2019bmn} grouped start-end pairs to build diverse action proposals, then scored them for final localization predictions.
BSN++~\cite{su2021bsn++} pointed out the imbalance problem over temporal scales of actions based on BSN.
Recently, ActionFormer~\cite{zhang2022actionformer} and TriDet~\cite{shi2023tridet} deployed transformer-based encoder as multi-scale backbone network, and BRN~\cite{kim2024boundary} resolved the issue of multi-scale features for TAD.

\begin{figure*}[t]
\centering
\includegraphics[width=17.70cm]{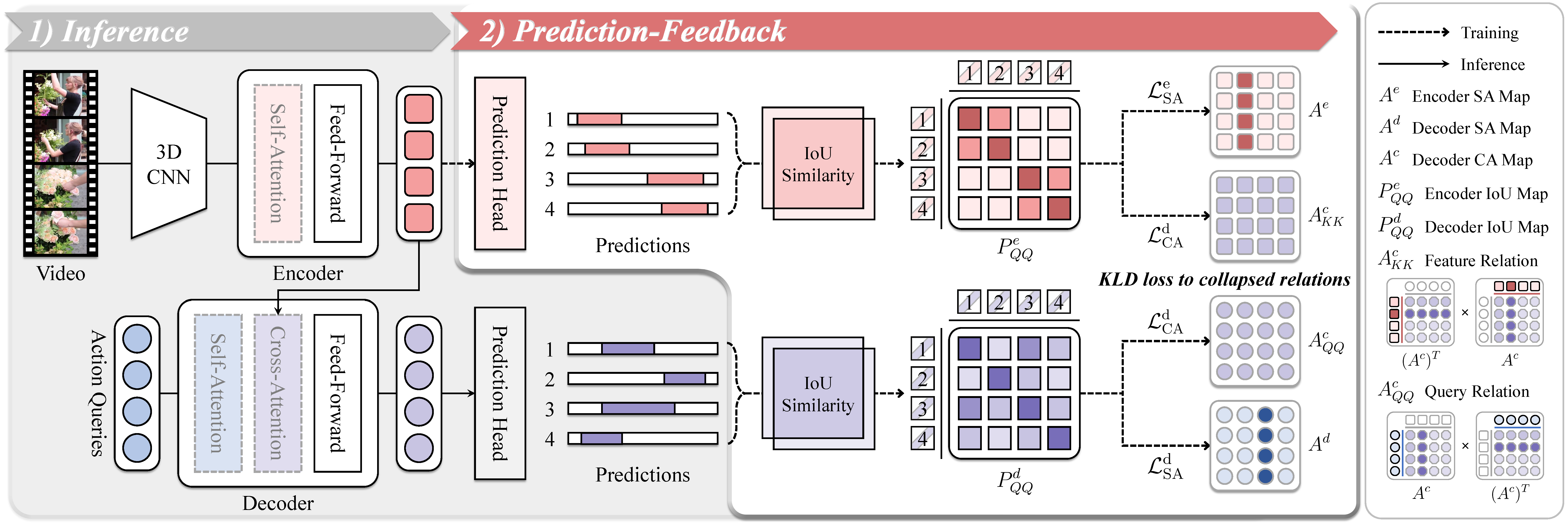}
\caption{\textbf{Overall architecture of the proposed framework, Pred-DETR.} 
The figure illustrates the entire framework of our model, Pred-DETR.
Pred-DETR consists of the two main parts: DETR architecture and prediction-feedback.
The encoder and decoder predictions are converted to the relation of Intersection-over-Union (IoU).
Then these IoU maps are utilized for prediction-feedback for the collapsed self- and cross-attention.
Note that the encoder predictions are deployed only for training.
}
\label{fig:architecture}
\end{figure*}

\subsection{DETR}
DETR~\cite{carion2020detr} is the first work to view object detection as a direct set prediction problem, and allow for end-to-end detection without any human heuristics such as non-maximum-suppression (NMS).
However, DETR demands 10 times longer training than the conventional approaches as bipartite matching is hard to optimize.
For this issue, Deformable DETR~\cite{xizhou2021deformable_detr} introduced sparse attention, which attends only a part of elements by learning to specify positions to focus on.
The subsequent DETR-based models~\cite{meng2021conditional_detr, shilong2022dab_detr} further advanced query representations through explicitly encoding box information, which effectively helps to stabilize training.

In TAD, the DETR-based methods are also deployed as DETR has reached a new state-of-the-art performance in object detection.
RTD-Net~\cite{tan2021relaxed} identified the problem of the dense attention in the encoder of DETR, which exhibits nearly uniform distribution causing that the self-attention layers act like an over-smoothing effect.
TadTR~\cite{liu2022tadtr} devised temporal deformable attention inspired by Deformable DETR~\cite{xizhou2021deformable_detr}.
ReAct~\cite{shi2022react} developed a new relation matching to enforce high correlation between queries with low-overlap and high feature similarity.
Also, LTP~\cite{kim2024long} proposed a pre-training strategy tailored for DETR.

Recently, Self-DETR~\cite{kim2023self} revealed the problem of the degraded DETR performance for TAD as attention collapse in the self-attention and proposed self-feedback to utilize a guidance map from the cross-attention maps for the self-attention modules.
Although it remarkably reduced the degree of the attention collapse, its optimal performance depends on the assumption that cross-attention is sound.
However, we discover that cross-attention has collapsed, and therefore introduce prediction-guided feedback which activates the cross-attention as well as the self-attention based on guidance from prediction relations.

%% file: sections/3_our_approach.tex
\section{Our Approach}
\label{sec:our_approach}
This section introduces our proposed method, prediction feedback for Pred-DETR.
To be specific, we first elaborate on the preliminaries, and discuss the attention collapse and predictions.
Then the explanation of our prediction-feedback mechanisms is followed depicting the overall framework in Fig.~\ref{fig:architecture}.
Moreover, we provide an extension of prediction-feedback to the encoder via the recent DETR architecture, only for training.
Finally, we summarize the overall objectives for Pred-DETR.

\subsection{Preliminary}
\label{sec:preliminary}
\noindent\textbf{DETR.} 
DETR~\cite{carion2020detr} adopts the transformer~\cite{vaswani2017attention} architecture and composed of two main components: encoder and decoder.
First, the encoder captures the global relationships among input features, which is achieved through similarity calculation of SA.

On the other hand, the decoder performs cross-attention operations between object queries and encoder features.
Here, object queries are learnable embedding vectors that learn positional information similar to anchors.
This mechanism ensures that each query attends to the most relevant parts of the input features processed by the encoder.

\vspace{3pt}
\noindent\textbf{Attention Mechanism.}
An attention module takes three inputs, projecting each into three latent spaces through linear layers.
The resulting projections are referred to as query $Q$, key $K$ and value $V$, respectively.
The attention map is then computed by matrix multiplication of $Q$ with the transpose of $K$, followed by applying the softmax activation function, scoring similarity between $Q$ and $K$.
By pooling $V$ with the scores followed by a linear projection, we obtain the output of the attention modules.
Formally, $Q$, $K$, and $V$ are represented as $\mathds{R}^{N_q \times D}$, $\mathds{R}^{N_k \times D}$, and $\mathds{R}^{N_v \times D}$, respectively, where $N_q$, $N_k$, and $N_v$ denote the lengths of $Q$, $K$ and $V$ while $D$ represents the number of channels.
When $Q$, $K$ and $V$ all have the same number of channels, the attention mechanism can be formulated as follows:
\begin{equation}
    \begin{aligned}
    \text{Attention}(Q, K, V)=AV; 
    A=\text{softmax}(\frac{QK^{\top}}{\sqrt{D}}),
    \label{eq:attention}
    \end{aligned}
\end{equation}
where $A\in{\mathds{R}^{N_q \times N_k}}$ is the attention map, $A^{\top}$ is the transpose of $A$.
For the SA module, the inputs to $Q$, $K$ and $V$ are the same while $Q$ is obtained from the object queries, while $K$ and $V$ are from the encoder features in the CA module. 

\vspace{3pt}
\noindent\textbf{DETR for TAD.}
There are three different things from original DETR for object detection.
First of all, we utilize video features from 3D CNN pre-trained on Kinetics~\cite{kay2017kinetics}.
Note that 3D CNN is frozen and only the temporal dimension is left for video features by global average pooling over the spatial dimensions.
Secondly, decoder queries act as action queries instead of object queries since decoder's outputs are used to predict the temporal action detection results. 
Lastly, DAB-DETR~\cite{shilong2022dab_detr} is adopted, consistent with Self-DETR~\cite{kim2023self}.

\noindent\textbf{Self-DETR.}
It is the first work to identify the collapse of encoder and decoder SA maps in DETR when applied to TAD.
To guide the collapsed SA maps, they process the CA map $A^{c}\in \mathds{R}^{N^{d}_{q}\times{N^{d}_{k}}}$ as follows:
\begin{equation}
\begin{aligned}
A^c_{QQ} = A^c \times (A{^c})^T,\ A^c_{KK} = (A^c)^T \times A^c,
\end{aligned}
\end{equation}
where $A^c_{QQ}\in \mathds{R}^{N^{d}_{q}\times{N^{d}_{q}}}$ and $A^c_{KK}\in \mathds{R}^{N^{d}_{k}\times{N^{d}_{k}}}$ indicate relations between queries and between keys, respectively.
In the next step, they ensure the encoder and decoder SA maps resemble $A^c_{KK}$ and $A^c_{QQ}$ by applying Kullback–Leibler (KL) divergence loss.
Please refer to the original paper for additional details.

\subsection{Prediction-Feedback}

\begin{figure*}[t]
\centering
\includegraphics[width=17.70cm]{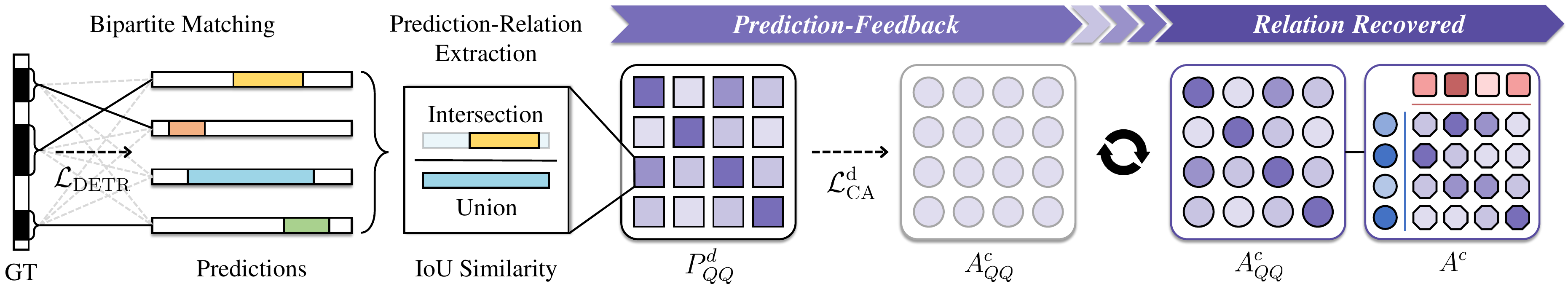}
\caption{\textbf{Prediction-Feedback.} 
This illustrates the detailed mechanism of prediction-feedback for the cross-attention.
The DETR predictions are diverse thanks to the bipartite matching.
By aligning attention with the IoU relation from the predictions, the query relation is recovered, alleviating the attention collapse.
}
\label{fig:feedback}
\end{figure*}

\noindent\textbf{Attention Collapse.}
The attention collapse problem is a phenomenon where the attention matrix becomes a rank-1 matrix to skip the attention module to prevent degeneration of the learning~\cite{dong2021rank_collapse}.
The collapsed attention outputs uniform values for all queries, resulting in the input being conveyed without additional representations through the residual connection.
In this paper, we newly discover the collapse of CA.
This problem brings a question for the assumption of the previous work that the CA is sound.
As a result, the complete remedy for the collapse through the entire attention modules is required.

\vspace{3pt}
\noindent\textbf{Feedback from Predictions.}
DETR is the first work of end-to-end detection mechanism without anchor boxes or non-maximum suppression (NMS).
For this, it uses learnable queries and bipartite matching to assign detection targets to queries since there is no pre-defined matches between predictions and the ground-truth.
As the matching is one-to-one mapping, the DETR predictions will be diverse because one query gets a negative loss when two queries produce similar localization results.
From this property, the guidance from the predictions can activate the collapse attention modules.

\vspace{3pt}
\noindent\textbf{Feedback for Cross-Attention.}
The bottom line is that both the prediction and the CA map are represented as the relations of the decoder queries to bridge between them.
Thereafter, we make the collapsed relation follow the diverse relation from predictions through an auxiliary objective.
The CA of the decoder connects the decoder queries with the encoder features to predict actions of interest.
One can regularize the collapsed CA map directly; however, to maintain the flexibility of attention results, we propose to guide the query relation $A^c_{QQ}$ extracted from the $A^c$.
The query relation is simply extracted from the CA, indicating how they attend to a similar group of encoder features.
Here, the purpose of reformulation of CA into self-relation is the opposite where Self-DETR utilizes $A^c_{QQ}$ as the guidance of feedback.

Subsequently, we design a guidance map based on predictions.
The key point here is that the query relation can also be extracted from the IoU similarity among predictions as depicted in Fig.~\ref{fig:feedback}.
To be specific, each prediction is identical to the refined decoder query and includes the time interval $t_i = \{s_i, e_i\}$ where $s_i$ and $e_i$ are the start and end times, respectively.
Therefore, the query relation is obtained from predictions by constructing an IoU similarity matrix $P^d_{QQ} \in \mathds{R}^{N^d_q \times N^d_q}$, where $N^d_q$ denotes the number of decoder queries, as below.
\begin{equation}
    \begin{aligned}
     P^d_{QQ}(i,j) = \frac{\max(0, \min(e_i, e_j) - \max(s_i, s_j))}{\max(e_i, e_j) - \min(s_i, s_j)},
    \label{eq:IoU}
    \end{aligned}
\end{equation}
where $i, j = 1, 2, \dots, N^d_q$.

After normalizing $P^d_{QQ}$ by a softmax function,
feedback objective with predictions for decoder CA is finally defined as follows:
\begin{equation}
    \begin{aligned}
    \mathcal{L}_{\text{CA}}^{\text{d}} = D_{KL}(A^c_{QQ}~||~P^d_{QQ}),
    \label{eq:QKQ}
    \end{aligned}
\end{equation}
where $D_{KL}$ is the KL divergence loss.

\vspace{3pt}
\noindent\textbf{Feedback for Decoder Self-Attention.}
Furthermore, we propose to guide the collapsed decoder SA maps with $P^d_{QQ}$.
Previous work has already shown the collapsed of the decoder SA maps and the positive impact of feedback on their recovery.
In addition, we enhance the feedback mechanism by utilizing $P^d_{QQ}$, which guarantees higher diversity than CA-based guidance.
Prediction-feedback objective for the decoder SA maps $A^d \in \mathds{R}^{N^d_q \times N^d_q}$ is defined as follows:
\begin{equation}
    \begin{aligned}
    \mathcal{L}_{\text{SA}}^{\text{d}} = D_{KL}(A^d~||~P^d_{QQ}).
    \label{eq:QQ}
    \end{aligned}
\end{equation}

\vspace{3pt}
\noindent\textbf{Feedback for Encoder Self-Attention.}
Besides the decoder, the encoder SA also suffers from a severe attention collapse.
Thanks to the query initialization from the encoder proposed in \cite{xizhou2021deformable_detr}, we obtain the predictions from the encoder.
Concretely, we add a linear layer on top of the encoder to predict, utilizing each encoder feature as action queries.
This allows us to construct the IoU relation between encoder features just as the IoU relation between decoder queries.
Accordingly, we devise the feedback objective for the encoder to follow the IoU map.

Similar to Eq.~\ref{eq:IoU} in the decoder SA, we denote the IoU matrix $P^e_{QQ} \in \mathds{R}^{N^e_q \times N^e_q}$ from the predictions of the encoder features, where $N^e_q$ is the number of the encoder features. After normalizing $P^e_{QQ}$ by a softmax function, the feedback objective with predictions for encoder SA is defined as
\begin{equation}
    \begin{aligned}
    \mathcal{L}_{\text{SA}}^{\text{e}} = D_{KL}(A^e~||~P^e_{QQ}).
    \label{eq:KK}
    \end{aligned}
\end{equation}

On the other hand, the CA map contains not only the query relation but also the relation of the encoder features. As such, we extend the prediction-feedback for the CA as described in Eq.~\ref{eq:QKQ}.
Specifically, the feature relation $A^c_{KK}$ is extracted from the CA by a matrix multiplication, similarly to $A^c_{QQ}$.
Here, $A^c_{KK}$ represents the similarities between encoder features to which similar groups of decoder queries attend. 
To conclude, $\mathcal{L}_{\text{CA}}^{\text{d}}$ defined in Eq.~\ref{eq:QKQ} is strengthened to
\begin{equation}
    \begin{aligned}
    \mathcal{L}_{\text{CA}}^{\text{d}} = D_{KL}(A^c_{QQ}~||~P^d_{QQ}) + D_{KL}(A^c_{KK}~||~P^e_{QQ}).
    \label{eq:QK}
    \end{aligned}
\end{equation}

\vspace{3pt}
\noindent\textbf{Discussion.}
During the initial training phases, the model generates undertrained predictions.
One might be concerned that the early feedback harms the learning of the model.
However, in the initial iterations, the objective of TAD is primarily optimized over the feedback, ensuring that undertrained feedback does not disrupt the training.
Additionally, note that the guidance derived from predictions does not constitute the optimal relation for attention.
The feedback acts as a regularizer, helping the attention maps stay close to the predictions and maintain their balance with the main objective.
Meanwhile, when the prediction-feedback relieves the collapse, the soundness of the CA is restored.
This brings about the restoration of the full functionality of the previous work, Self-DETR.
Experimental results demonstrate that the recovered CA remarkably boosts its performance.

\subsection{Objectives}
\noindent\textbf{DETR.}
Let us denote the ground-truths, and the $M$ predictions as $y$, $\hat{y}={\hat{y_i}_{i=1}^{M}}$, respectively.
For the bipartite matching between the ground-truth and prediction sets,
the optimal matching is defined to search for the permutation of $M$ elements $j \in J_M$ with the minimal cost as below:
\begin{equation}
    \begin{aligned}
    \hat{j} = \argmin_{j \in J_M} \sum_{i}^{M}\mathcal{L}_{\text{match}}(y_i, \hat{y}_{j(i)}),
    \label{eq:bipartite_matching}
    \end{aligned}
\end{equation}
where $L_{\text{match}(y_i, \hat{y}_{j(i)})}$ is a pair-wise matching cost between $y_i$ and the prediction with the index from $j(i)$, which produces the index $i$ from the permutation $j$.

Next, we denote each ground-truth action as $y_i = (c_i, t_i)$, where $c_i$ is the target class label with the background category $\emptyset$ and $t_i$ is the time intervals of the start and end times.
For the prediction with the index $j(i)$, we define the probability of the class $c_i$ as $\hat{p}_{j(i)}(c_i)$ and the predictions of the time intervals as $\hat{t}_{\hat{j}(i)}$.
Then $\mathcal{L}_{\text{match}}(y_i, \hat{y}_{j(i)})$ is defined as below:
\begin{equation*}
    \begin{aligned}
    \mathcal{L}_{\text{match}}(y_i, \hat{y}_{j(i)}) = -\mathbb{1}_{c_i \neq \emptyset}~\hat{p}_{j(i)}(c_i) + \mathbb{1}_{c_i \neq \emptyset}~\mathcal{L}_{\text{reg}}(t_i, \hat{t}_{j(i)}),
    \label{eq:matching_cost}
    \end{aligned}
\end{equation*}
where $\mathcal{L}_{\text{reg}}(t_i, \hat{t}_{j(i)})$ is the regression loss between the ground-truth $t_i$ and the prediction $\hat{t}$ with the index $j(i)$.
The regression loss $\mathcal{L}_{\text{reg}}$ consists of L1 and Interaction-over-Union (IoU) losses as in the DETR-based methods.
Finally, we formulate the main objective as follows:
\begingroup
\small{
\begin{equation}
    \begin{aligned}
    \mathcal{L}_{\text{DETR}}(y, \hat{y}) = \sum_{i=1}^{M}[-\log{\hat{p}_{\hat{j}(i)}(c_i)} + \mathbb{1}_{c_i \neq \emptyset}\mathcal{L}_{\text{reg}}(t_i, \hat{t}_{\hat{j}(i)})],
    \label{eq:detr_loss}
    \end{aligned}
\end{equation}
}
\endgroup
where $\hat{j}$ is the optimal assignment from Eq.~\ref{eq:bipartite_matching}.

\vspace{3pt}
\noindent\textbf{Full Objectives.}
To summarize the objectives for our framework, Pred-DETR,
the full objective is can be described as below:
\begin{equation}
    \begin{aligned}
    \mathcal{L} = \mathcal{L}_{\text{DETR}} + \lambda^e_{\text{SA}} \mathcal{L}_{\text{SA}}^{\text{e}} + \lambda^d_{\text{SA}} \mathcal{L}_{\text{SA}}^{\text{d}} + \lambda^d_{\text{CA}} \mathcal{L}_{\text{CA}}^{\text{d}},
    \label{eq:full_objective}
    \end{aligned}
\end{equation}
where $\lambda^e_{\text{SA}}$, $\lambda^d_{\text{SA}}$ and $\lambda^d_{\text{CA}}$ are the weights for the prediction-feedback losses for the encoder and decoder.

%% file: sections/4_experiments.tex
\section{Experiments}
\subsection{Datasets}
In this paper, we utilize four challenging benchmarks of temporal action detection: THUMOS14~\cite{jiang2014thumos14}, ActivityNet-v1.3~\cite{caba2015activitynet}, HACS~\cite{zhao2019hacs} and FineAction~\cite{liu2022fineaction}.
\textbf{THUMOS14} has 200 and 213 videos for the training and validation sets, respectively. 
The dataset has 20 action classes related to sports. 
\textbf{ActivityNet-v1.3} contains 19,994 videos with 200 action classes. 10024, 4926, and 5044 videos are for training, validation, and testing, respectively.
\textbf{HACS} contains 37613 and 5981 videos are for training, validation, respectively, with 200 classes, shared by ActivityNet-v1.3.
\textbf{FineAction} contains daily events with 106 categories and 16732 videos.
THUMOS14 and FineAction contain many short actions while a majority of videos in ActivityNet-v1.3 and HACS have long actions.

\subsection{Implementation Details}
In this section, we briefly deliver the implementation details.
For the detailed description, we recommend to refer to the supplementary materials.

\vspace{3pt}
\noindent\textbf{Architecture.}
We utilize the features of I3D~\cite{carreira2017i3d} pre-trained on Kinetics~\cite{kay2017kinetics} for THUMOS14 and ActivityNet-v1.3. 
Also, we adopt SlowFast~\cite{feichtenhofer2019slowfast} and VideoMAEv2-g~\cite{wang2023videomaev2} for HACS and FineAction, respectively.
We are based on a temporal version of DAB-DETR as in Self-DETR.
The number of layers of the encoder and decoder is $2$, and $4$, respectively.
The number of the queries is $40$.
We set the weights $\lambda^e_{\text{SA}}$, $\lambda^d_{\text{SA}}$ and , $\lambda^d_{\text{CA}}$ of the losses of the prediction-feedback for the encoder and decoder as $2$.

\begingroup
\setlength{\tabcolsep}{2.30pt} 
\renewcommand{\arraystretch}{1.0} 
\begin{table*}[t]
\centering
\begin{tabular}{l||c|ccccc|c||c|ccc|c}
    \hline\hline
    \multirow{2}{*}{Method} &
    \multicolumn{7}{c||}{THUMOS14} & \multicolumn{5}{c}{ActivityNet-v1.3} \\
    \cline{2-13}
    & Feat. & $0.3$ & $0.4$ & $0.5$ & $0.6$ & $0.7$ & Avg. & Feat. & $0.5$ & $0.75$ & $0.95$ & Avg. \\ 
    \hline\hline
    \rowcolor{gray!25}\multicolumn{13}{l}{\textit{\textbf{Standard Methods}}} \\
    \hline\hline
    BMN~\cite{lin2019bmn} & TSN & $56.0$ & $47.4$ & $38.8$ & $29.7$ & $20.5$ & $38.5$ & TSN & $50.07$ & $34.78$ & $8.29$ & $33.85$ \\
    G-TAD~\cite{xu2020gtad} & TSN & $54.5$ & $47.6$ & $40.2$ & $30.8$ & $23.4$ & $39.3$ & TSN & $50.36$ & $34.60$ & $9.02$ & $34.09$ \\
    AFSD~\cite{lin2021salient} & I3D & $67.3$ & $62.4$ & $55.5$ & $43.7$ & $31.1$ & $52.0$ & I3D & $52.40$ & $35.30$ & $6.50$ & $34.40$ \\
    TAGS~\cite{nag2022tags} & I3D & $68.6$ & $63.8$ & $57.0$ & $46.3$ & $31.8$ & $52.8$ & I3D & $56.30$ & $36.80$ & $\boldsymbol{9.60}$ & $36.50$ \\
    ActionFormer~\cite{zhang2022actionformer} & I3D & $82.1$ & $77.8$ & $71.0$ & $59.4$ & $43.9$ & $66.8$ & I3D & $53.50$ & $36.20$ & $8.20$ & $35.60$ \\
    TriDet~\cite{shi2023tridet} & I3D & $83.6$ & $\boldsymbol{80.1}$ & $\boldsymbol{72.9}$ & $\boldsymbol{62.4}$ & $47.4$ & $\boldsymbol{69.3}$ & TSP & $54.70$ & $38.00$ & $8.40$ & $36.80$ \\
    DyFaDet~\cite{yang2025dyfadet} & I3D & $\boldsymbol{84.0}$ & $\boldsymbol{80.1}$ & $72.7$ & $61.1$ & $\boldsymbol{47.9}$ & $69.2$ & TSP & $\boldsymbol{58.10}$ & $\boldsymbol{39.60}$ & $8.40$ & $\boldsymbol{38.50}^*$ \\
    \hline\hline
    \rowcolor{gray!25}\multicolumn{13}{l}{\textit{\textbf{DETR-based Methods}}} \\
    \hline\hline
    RTD-Net~\cite{tan2021relaxed} & I3D & $68.3$ & $62.3$ & $51.9$ & $38.8$ & $23.7$ & $49.0$ & I3D & $47.21$ & $30.68$ & $8.61$ & $30.83$ \\
    TadTR~\cite{liu2022tadtr} & I3D & $74.8$ & $69.1$ & $60.1$ & $46.6$ & $32.8$ & $56.7$ & I3D & $52.83$ & $37.05$ & $\boldsymbol{10.83}$ & $36.11$ \\
    ReAct~\cite{shi2022react} & TSN & $69.2$ & $65.0$ & $57.1$ & $47.8$ & $35.6$ & $55.0$ & TSN & $49.60$ & $33.00$ & $8.60$ & $32.60$ \\
    Self-DETR~\cite{kim2023self} & I3D & $74.6$ & $69.5$ & $60.0$ & $47.6$ & $31.8$ & $56.7$ & I3D & $52.25$ & $33.67$ & $8.40$ & $33.76$ \\
    \hline
    Pred-DETR (Ours) & I3D & $80.0$ & $73.5$ & $64.6$ & $52.3$ & $37.6$ & $61.6$ & I3D & $54.17$ & $36.43$ & $9.53$ & $36.00$ \\
    Pred-DETR (Ours) & VM2 & $\boldsymbol{84.1}$ & $\boldsymbol{80.0}$ & $\boldsymbol{72.2}$ & $\boldsymbol{60.4}$ & $\boldsymbol{45.8}$ & $\boldsymbol{68.5}$ & I3D & $\boldsymbol{58.38}$ & $\boldsymbol{39.14}$ & $9.92$ & $\boldsymbol{38.62}^*$ \\
    \hline\hline
\end{tabular}
\caption{\textbf{The comparison results with the state-of-the-art on THUMOS14 and ActivityNet-v1.3.} 
`Feat.' indicates the backbone features including I3D, TSN, TSP (R(2+1)D) and VM2.
`*' means using UniFormer-v2 classification on ActivityNet-v1.3.
}
\label{tab:main}
\end{table*}
\endgroup

\begingroup
\setlength{\tabcolsep}{1.10pt} 
\renewcommand{\arraystretch}{1.0} 
\begin{table}[t]
\centering
\begin{tabular}{l|c||ccc|c}
    \hline\hline
    Method & Feat. & $0.5$ & $0.75$ & $0.95$ & Avg. \\
    \hline\hline
    \rowcolor{gray!25}\multicolumn{6}{l}{\textit{\textbf{Standard Methods}}} \\
    \hline\hline
    G-TAD~\cite{xu2020gtad} & I3D & $41.1$ & $27.6$ & $88.3$ & $27.5$ \\
    BMN~\cite{lin2019bmn} & SF & $52.5$ & $36.4$ & $10.4$ & $35.8$ \\
    TCANet~\cite{qing2021temporal} & SF & $54.1$ & $37.2$ & $11.3$ & $36.8$ \\
    TriDet~\cite{shi2023tridet} & SF & $\boldsymbol{56.7}$ & $\boldsymbol{39.3}$ & $\boldsymbol{11.7}$ & $\boldsymbol{38.6}$ \\
    \hline\hline
    \rowcolor{gray!25}\multicolumn{6}{l}{\textit{\textbf{DETR-based Methods}}} \\
    \hline\hline
    TadTR~\cite{liu2022tadtr} & I3D & $47.1$ & $32.1$ & $10.9$ & $32.1$ \\
    Self-DETR$^\dagger$~(Kim et al. 2023) & I3D & $49.9$ & $31.1$ & $9.3$ & $31.8$ \\
    \hline
    Pred-DETR~(Ours) & I3D & $51.5$ & $32.7$ & $10.2$ & $33.3$ \\
    Pred-DETR~(Ours) & SF & $\boldsymbol{56.5}$ & $\boldsymbol{36.8}$ & $\boldsymbol{12.1}$ & $\boldsymbol{37.4}$ \\
    \hline\hline
\end{tabular}
\caption{\textbf{The comparison results on the HACS dataset.} 
`$\dagger$' indicates our reproduced version.
}
\label{tab:hacs_results}
\end{table}
\endgroup

\begingroup
\setlength{\tabcolsep}{1.80pt} 
\renewcommand{\arraystretch}{1.0} 
\begin{table}[t]
\centering
\small
\begin{tabular}{l|c||ccc|c}
    \hline\hline
    Method & Feat. & $0.5$ & $0.75$ & $0.95$ & Avg. \\
    \hline\hline
    \rowcolor{gray!25}\multicolumn{6}{l}{\textit{\textbf{Standard Methods}}} \\
    \hline\hline
    DBG~\cite{lin2020fast} & I3D & $10.7$ & $6.4$ & $2.5$ & $6.8$ \\
    BMN~\cite{lin2019bmn} & I3D & $13.7$ & $8.8$ & $3.1$ & $9.1$ \\
    G-TAD~\cite{xu2020gtad} & I3D & $14.4$ & $8.9$ & $3.1$ & $9.3$ \\
    ActionFormer~(Zhang et al. 2022) & VM2 & $\boldsymbol{29.1}$ & $\boldsymbol{17.7}$ & $\boldsymbol{5.1}$ & $\boldsymbol{18.2}$ \\
    \hline\hline
    \rowcolor{gray!25}\multicolumn{6}{l}{\textit{\textbf{DETR-based Methods}}} \\
    \hline\hline
    TadTR$^\dagger$~\cite{liu2022tadtr} & VM2 & $21.3$ & $8.7$ & $0.4$ & $10.3$ \\
    Self-DETR$^\dagger$~(Kim et al. 2023) & VM2 & $22.1$ & $9.2$ & $0.4$ & $10.7$ \\
    \hline
    Pred-DETR~(Ours) & VM2 & $\boldsymbol{25.9}$ & $\boldsymbol{12.2}$ & $\boldsymbol{1.0}$ & $\boldsymbol{13.4}$ \\
    \hline\hline
\end{tabular}
\caption{\textbf{The comparison results on the FineAction dataset.} 
`$\dagger$' indicates our reproduced version.
}
\label{tab:fineaction_results}
\end{table}
\endgroup

\vspace{3pt}
\noindent\textbf{Enhanced DAB-DETR for TAD.} 
We also introduce advanced tricks on DAB-DETR including stable matching~\cite{liu2023stable}, hybrid matching~\cite{jia2023hybrid} and the two-stage mechanism from Deformable-DETR.
Stable matching utilizes the IoU value between the prediction and the ground-truth as the target value of the class probability.
It is closely related to the actionness regression in TadTR.
Note that we do not utilize the predictions from the encoder as the initial decoder queries.
We found that stable matching remarkably improves the performance, aligned with the results of TadTR.
However, the two-stage mechanism slightly improves it because it is introduced for the prediction-feedback.
We also report a study for the benefits from each component in the supplementary materials.

\subsection{Main Results}
\noindent\textbf{Comparison with the State-of-the-Art.} 
Table.~\ref{tab:main} shows the comparison results with the state-of-the-art methods on THUMOS14 and ActivityNet-v1.3.
Furthermore, Table.~\ref{tab:hacs_results} and Table.~\ref{tab:fineaction_results} show the comparison results on HACS and FineAction.
Pred-DETR outperforms DETR-based methods over various benchmarks.
The first section denoted by `Standard Methods' contains non-DETR approaches, and the second one includes DETR-based models.
Also, in the DETR-based models, RTD-Net, Self-DETR, and ours are based on standard attention.
Also, TadTR and ReAct are based on deformable attention.
We also indicate the backbone features by `Feats'.
Most approaches utilize the TSN~\cite{wang2016tsn} or I3D features while some methods also adopt the TSP~\cite{alwassel2021tsp} features.

In the table, our model remarkably outperforms all DETR-based models on all benchmarks.
It demonstrates that when the attention collapse is relieved, the original-DETR architecture can be comparable or superior to the Deformable-DETR architecture in TAD, aligned with the observation in object detection~\cite{lin2023plain-detr}.
More interestingly, Pred-DETR best performs on ActivityNet-v1.3 including the non-DETR methods.
DETR-based methods tend to produce better results on ActivityNet and HACS than those on THUMOS14 and FineAction.
This could be because ActivityNet and HACS mostly contain long actions while THUMOS14 and FineAction include many short instances.
Predicting short actions precisely requires high temporal resolution while DETR does not handle yet such a long sequence due to the query-based architecture.
Nevertheless, recent DETR models including ours show superior performances over the non-DETR models which handle short-length sequences except for ActionFormer and TriDet.

\subsection{Analysis}

\begin{figure}[t]
\centering
\includegraphics[width=8.35cm]{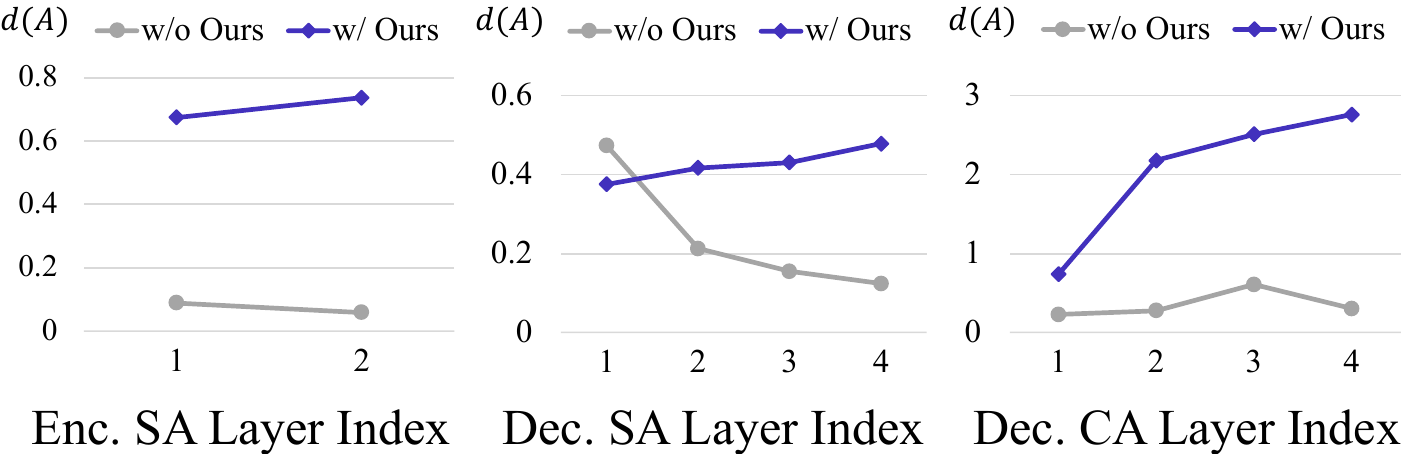}
\caption{\textbf{Diversity of attention maps.}
Diversity for cross- and self-attention for test samples of ActivityNet-v1.3.
}
\label{fig:diversity}
\end{figure}

\begin{figure}[t]
\centering
\includegraphics[width=8.40cm]{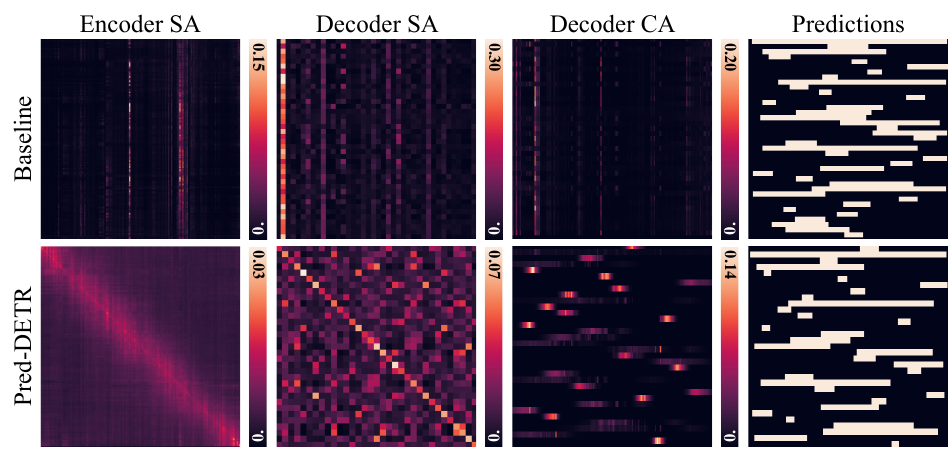}
\caption{\textbf{Attention maps.}
The figure shows self- and cross-attention maps from a test sample in ActivityNet-v1.3.
}
\label{fig:attention}
\end{figure}

\begingroup
\setlength{\tabcolsep}{1.90pt} 
\renewcommand{\arraystretch}{1.0} 
\begin{table}[t]
\centering
\small
\begin{tabular}{c|c|c||ccc|c||ccc|c}
    \hline\hline
    \multirow{2}{*}{$\mathcal{L}_{\text{SA}}^{\text{e}}$} & \multirow{2}{*}{$\mathcal{L}_{\text{SA}}^{\text{d}}$} & \multirow{2}{*}{$\mathcal{L}_{\text{CA}}^{\text{d}}$} &
    \multicolumn{4}{c||}{THUMOS14} & \multicolumn{4}{c}{ActivityNet-v1.3} \\
    \cline{4-11}
    & & & $0.3$ & $0.5$ & $0.7$ & Avg. & $0.5$ & $0.75$ & $0.95$ & Avg. \\
    \hline\hline
    $\cdot$ & $\cdot$ & $\cdot$ & $73.8$ & $57.0$ & $26.5$ & $53.5$ & $52.65$ & $33.79$ & $8.93$ & $34.14$ \\
    \checkmark & $\cdot$ & $\cdot$ & $79.7$ & $64.0$ & $32.0$ & $59.7$ & $56.91$ & $36.27$ & $9.59$ & $36.67$ \\
    $\cdot$ & \checkmark & $\cdot$ & $77.0$ & $61.9$ & $34.6$ & $58.8$ & $54.60$ & $34.65$ & $9.25$ & $35.11$ \\
    $\cdot$ & $\cdot$ & \checkmark & $79.0$ & $64.1$ & $36.1$ & $60.8$ & $57.59$ & $38.64$ & $9.81$ & $38.14$ \\
    \checkmark & \checkmark & $\cdot$ & $79.0$ & $63.3$ & $35.4$ & $60.5$ & $57.63$ & $37.88$ & $9.46$ & $37.65$ \\
    \checkmark & \checkmark & \checkmark & $80.0$ & $64.6$ & $37.6$ & $61.6$ & $58.38$ & $39.14$ & $9.92$ & $38.62$ \\
    \hline\hline
\end{tabular}
\caption{\textbf{Ablation on prediction-feedback.} 
The ablation study is conducted on THUMOS14 and ActivityNet-v1.3.
}
\label{tab:ablation_feedback}
\end{table}
\endgroup

\begingroup
\setlength{\tabcolsep}{2.80pt} 
\renewcommand{\arraystretch}{1.0} 
\begin{table}[t]
\centering
\begin{tabular}{l|l||ccc|c}
    \hline\hline
    Cross-Attn. & Self-Attn. & $0.5$ & $0.75$ & $0.95$ & Avg. \\
    \hline\hline
    - & From CA & $56.14$ & $36.10$ & $9.12$ & $36.18$ \\
    - & Pred Relation & $57.63$ & $37.88$ & $9.46$ & $37.65$ \\
    Pred Relation & From CA & $57.80$ & $38.57$ & $10.16$ & $38.32$ \\
    Pred Relation & Pred Relation & $58.38$ & $39.14$ & $9.92$ & $38.62$ \\
    \hline\hline
    Ground-Truth & Pred Relation & $53.41$ & $33.76$ & $8.96$ & $34.12$ \\
    Pred Intervals & Pred Relation & $53.41$ & $34.33$ & $8.93$ & $34.62$ \\
    Pred Relation & Pred Relation & $58.38$ & $39.14$ & $9.92$ & $38.62$ \\
    \hline\hline
\end{tabular}
\caption{\textbf{Prediction-feedback targets.} 
We conduct the study on the prediction-feedback targets for the cross- and self-attention on ActivityNet-v1.3.
}
\label{tab:feedback_targets}
\end{table}

\noindent\textbf{Diversity.} 
To clearly validate the effect of the feedback for the collapse, we measure the diversity of cross- and self-attention maps according to~\cite{dong2021rank_collapse, kim2023self}.
The diversity $d(A)$ for the attention map $A$ is the measure of the closeness between the attention map and a rank-1 matrix as below:
\begingroup
\small{
\begin{equation*}
    \begin{aligned}
    d(A) = \| A - \boldsymbol{1}a^\top \|~/~\| A \|, \text{where}~a = \argmin_{a^{\prime}}\| A - \boldsymbol{1}a^{\prime\top} \|,
    \label{eq:diversity}
    \end{aligned}
\end{equation*}
}
\endgroup
where $\| \cdot \|$ denotes the $\ell_1$,$\ell_\infty$-composite matrix norm, $a$, $a'$ are column vectors of the attention map $A$, and $\boldsymbol{1}$ is an all-ones vector.
Note that the rank of $\boldsymbol{1}a^\top$ is 1, and therefore, a smaller value of $d(A)$ means $A$ is closer to a rank-1 matrix.

Fig.~\ref{fig:diversity} shows the diversity on each layer of the encoder and decoder for the baseline DETR and Pred-DETR.
The diversity is measured on the test set on ActivityNet-v1.3 averaged over all test samples.
As the model depth gets deeper, the diversity of the baseline decreases close to $0$.
However, the diversity of Pred-DETR does not fall down and even increases.
From this, Pred-DETR effectively relieves the collapse.

\vspace{3pt}
\noindent\textbf{Attention Maps.} 
Fig.~\ref{fig:attention} shows the visualization of the self- and cross-attention maps from the encoder and decoder.
As shown, the baseline DETR exhibits the attention collapse in all attention modules.
On the other hand, our model does not suffer from the collapse and shows expressive attention.

\vspace{3pt}
\noindent\textbf{Ablation on Prediction-Feedback.} 
In order to validate the benefits from each component in our framework, we conducted the ablation study on the self-feedback objectives.
In Pred-DETR, we have three types of feedback for 1) decoder cross-attention ($\mathcal{L}_{\text{CA}}^{\text{d}}$), 2) decoder self-attention ($\mathcal{L}_{\text{SA}}^{\text{d}}$), and 3) encoder self-attention ($\mathcal{L}_{\text{SA}}^{\text{e}}$).

Table.~\ref{tab:ablation_feedback} shows the ablation results.
As shown, each type of prediction-feedback clearly improves the performance.
Also, when we introduce all three kinds of prediction-feedback, the benefits become the most.
The prediction-feedback for the cross-attention modules brings the most performance gain as they are the central part of DETR.

\vspace{3pt}
\noindent\textbf{Prediction-Feedback Targets.} 
\endgroup
As for the targets for the self-attention in feedback, we can also adopt the guidance map from the cross-attention proposed in Self-DETR~\cite{kim2023self}.
The upper of Table.~\ref{tab:feedback_targets} shows the results with Self-DETR.
When we do not use the prediction-feedback for the cross-attention, we can see that the the feedback from the predictions (denoted by `Pred Relation' in the table) show superior performance to that from the cross-attention (From CA).
Also, when Self-DETR is introduced with our prediction-feedback for the cross-attention, the performance gain becomes a way larger because the attention collapse of the cross-attention is remarkably relieved.

In our prediction-feedback for cross-attention, we propose to utilize the indirect relation from the cross-attention.
One may think a direct solution as the objective where the ground-truth or prediction intervals are matched to the cross-attention map.
However, we claim that this way significantly harms the diversity of the representations for the CA mainly because we do not exactly know where the cross-attention should focus on.
The bottom of Table.~\ref{tab:feedback_targets} shows the results with three types of the targets on ActivityNet-v1.3.
The targets from the intervals of the ground-truth or predictions (`Ground-Truth' and `Prediction Intervals', respectively) degrade the performance as expected.
However, the indirect way with the relation of the predictions (Prediction Relation) remarkably improves the performance.

%% file: sections/5_conclusion.tex
\section{Conclusion}
In this paper, we discovered the attention collapse in the cross-attention of DETR for TAD.
We found that the model exhibits a clearly different pattern from the predictions, which is a short-cut phenomenon from the collapse.
To this end, we proposed Prediction-Feedback DETR (Pred-DETR) to align the attention with the predictions.
By providing an auxiliary objective with the guidance from the predictions, the prediction-feedback remarkably relieved the degree of the collapse.
Our extensive experiments demonstrated that Pred-DETR recovered the diversity of the attention with the state-of-the-art performance over DETR models on THUMOS14, ActivityNet-v1.3, HACS, and FineAction.

%% file: sections/7_supplementary.tex
\renewcommand{\thepage}{A\arabic{page}}  
\renewcommand{\thesection}{A}   
\renewcommand{\thetable}{A\arabic{table}}   
\renewcommand{\thefigure}{A\arabic{figure}}

\setcounter{section}{0}
\setcounter{table}{0}
\setcounter{figure}{0}

\section{Additional Details}

\noindent\textbf{Architecture.}
We utilize the features of I3D pre-trained on Kinetics for THUMOS14 and ActivityNet-v1.3. 
Also, we adopt SlowFast and VideoMAEv2-g for HACS and FineAction, respectively.
We are based on a temporal version of DAB-DETR as in Self-DETR.
Note that size-modulated attention, learnable anchors and iterative updates in DAB-DETR are also used in TadTR and ReAct as they are based on Deformable-DETR.
The number of layers of the encoder and decoder is $2$, and $4$, respectively.
The number of the queries is $40$.
We set the weights $\lambda^e_{\text{SA}}$, $\lambda^d_{\text{SA}}$ and , $\lambda^d_{\text{CA}}$ of the losses of the prediction-feedback for the encoder and decoder as $2$.

\vspace{3pt}
\noindent\textbf{Training.}
As for all datasets, we use the AdamW~\cite{DBLP:conf/iclr/adamw} optimizer with the batch size of 16.
We use 192 length of temporal features for all benchmarks.
The learning rate decreases by cosine anealing for all benchmarks.
In THUMOS14 and FineAction, we train the framework for 30 epochs.
As for ActivityNet-v1.3 and HACS, 20 epochs are taken for training with a warm-up of 5 epochs. 
In addition, we resize the features of a video with linear interpolation to 192 length for ActivityNet-v1.3 and HACS.

\vspace{3pt}
\noindent\textbf{Inference.} 
We slice the temporal features with a 192-length window with overlap of 48 for THUMOS14 and FineAction. 
As for ActivityNet-v1.3 and HACS, we resize the features to 192 length as done in training.
Also, we use the top 100, 100, 200, 200 predictions after non-maximum suppression (NMS) for the final localization results for ActivityNet-v1.3, HACS, THUMOS14 and FineAction, respectively.
We use SoftNMS with the NMS threshold of $0.40$.
For the class label, we fuse our classification scores with the top-1 video-level predictions of an external classifier as done in the baselines for ActivityNet-v1.3 and HACS.

\vspace{3pt}
\noindent\textbf{Enhanced DAB-DETR for TAD.} 
In order to properly validate our benefits, we first examine and deliver the effect from the tricks.
Firstly, we employ stable matching~\cite{liu2023stable}, indicated by `Stable' in the table.
Shortly, stable matching utilizes the IoU value between the prediction and the ground-truth as the target value of the class probability.
It is closely related to the actionness regression in TadTR.
Secondly, we also introduce hybrid matching (Hybrid)~\cite{jia2023hybrid}.
Lastly, we deploy the two-stage mechanism from Deformable-DETR as mentioned in the method to make it possible to utilize the predictions from the encoder features.
Note that we do not utilize the predictions from the encoder as the initial decoder queries.

Table.~\ref{tab:enhanced_DABDETR} shows the effect of each component.
As shown, the stable matching remarkably improves the performance, aligned with the results of TadTR.
Moreover, hybrid matching further enhances the performance.
However, the two-stage mechanism slightly improves it because the two-stage structure is introduced for the prediction-feedback.
We found that the stable matching remarkably improves the performance, aligned with the results of TadTR.

\begingroup
\setlength{\tabcolsep}{2.25pt} 
\renewcommand{\arraystretch}{1.0} 
\begin{table}[t]
\centering
\begin{tabular}{c|c|c|c|c}
    \hline\hline
    Stable & Hybrid & Two-stage & 
    THUMOS14 & ActivityNet-v1.3 \\
    \hline\hline
    $\cdot$ & $\cdot$ & $\cdot$ & $50.3$ & $31.08$ \\
    \checkmark & $\cdot$ & $\cdot$ & $52.3$ & $33.15$ \\
    \checkmark & \checkmark & $\cdot$ & $53.1$ & $34.10$ \\
    \checkmark & \checkmark & \checkmark & $53.5$ & $34.14$ \\
    \hline\hline
\end{tabular}
\caption{\textbf{Enhanced DAB-DETR. } 
In this paper, we deploy advanced DETR techniques for TAD.
}
\label{tab:enhanced_DABDETR}
\end{table}
\endgroup

\vspace{3pt}
\noindent\textbf{Prediction-Feedback.}
In the main paper, we elaborate the prediction-feedback without consideration of the multiple layers for a clear explanation preventing from being overwhelmed by excessive details.
As the encoder and decoder have the multi-layer architecture, we deliver the specific implementation here.

First of all, as for the decoder cross-attention, we have two loss terms, $D_{KL}(A^c_{QQ}~||~P^d_{QQ})$ and $D_{KL}(A^c_{KK}~||~P^e_{QQ})$, as defined in Eq.~7 of the paper.
The first one is from the decoder predictions, and the second one is from the encoder predictions.
For the decoder predictions, we use the predictions from the last layer of the decoder as $P^d_{QQ}$.
Also, we average the CA maps from multiple layers as $A^c_{QQ}$.
Then we calculate $D_{KL}(A^c_{QQ}~||~P^d_{QQ})$ as the first term of the prediction-feedback for the CA.
Next, as for the encoder predictions, we first note that the encoder predictions have also multiple layers where the prediction head is shared as one layer, but the basis of temporal scales has four values.
The details of the two-stage architecture in DETR can be found in recent DETR models~\cite{xizhou2021deformable_detr, shilong2022dab_detr, zhang2022dino}.
For the multi-scale encoder predictions, we first interpolate the predictions so that the predictions from all scales have the same temporal length as the finest scale (the longest).
Then we average the predictions with the weights from the confidence scores of the predictions on the scale axis as $P^e_{QQ}$.
The CA maps are also averaged on the layer axis as $A^c_{KK}$
Finally, we calculate $D_{KL}(A^c_{KK}~||~P^e_{QQ})$ as the second term of the prediction-feedback for the CA.

As for the decoder self-attention, we calculate the loss defined in Eq.~5 of the paper for each layer.
Also, for the encoder self-attention, we can get $P^e_{QQ}$ in the same way for the decoder CA.
We average the self-attention maps on the layer axis to get $A^e$.
Then we calculate the objective for the prediction-feedback for the encoder SA as defined in Eq.~6 of the paper.

\section{Additional Results}

\begin{figure}[t]
\centering
\includegraphics[width=8.40cm]{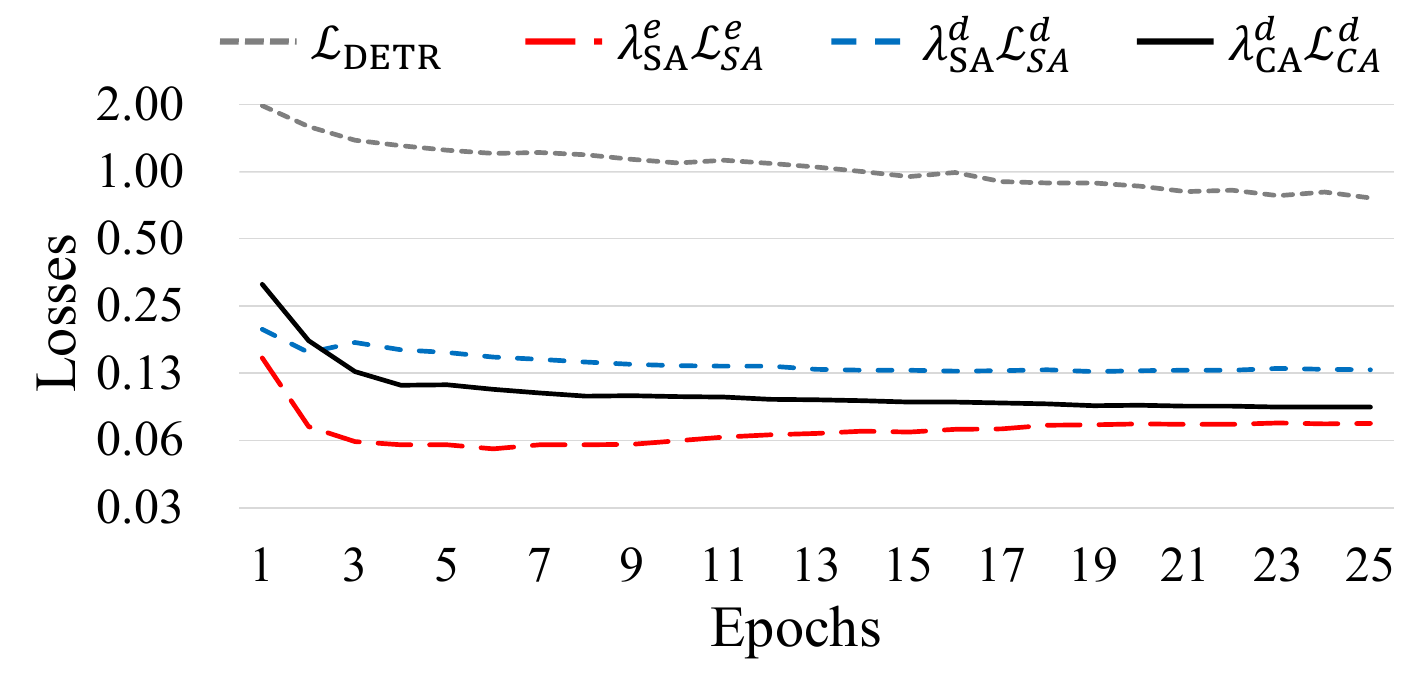}
\caption{\textbf{Feedback Losses.}
The figure shows the DETR and feedback losses on ActivityNet-v1.3.
}
\label{fig:losses}
\end{figure}

\begingroup
\setlength{\tabcolsep}{0.60pt} 
\renewcommand{\arraystretch}{1.0} 
\begin{table}[t]
\centering
\begin{tabular}{l|c||ccc|c}
    \hline\hline
    Method & Feat. & $0.3$ & $0.5$ & $0.7$ & Avg. \\
    \hline\hline
    \rowcolor{gray!25}\multicolumn{6}{l}{\textit{\textbf{Standard Methods}}} \\
    \hline\hline
    ActionFormer~(Zhang et al. 2022) & VM2 & $84.0$ & $73.0$ & $47.7$ & $69.6$ \\
    TriDet~\cite{shi2023tridet} & VM2 & $84.8$ & $73.3$ & $48.8$ & $70.1$ \\
    \hline\hline
    \rowcolor{gray!25}\multicolumn{6}{l}{\textit{\textbf{DETR-based Methods}}} \\
    \hline\hline
    Pred-DETR~(Ours) & VM2 & $84.1$ & $72.2$ & $45.8$ & $68.5$ \\
    \hline\hline
\end{tabular}
\caption{\textbf{The THUMOS14 results with VM2 features.} 
VM2 indicates VideoMAEv2-g features.
}
\label{tab:thumos14_vm2}
\end{table}
\endgroup

\noindent\textbf{Feedback Losses.} 
As mentioned in the discussion of the method section, the undertrained model produces noisy predictions in the early steps of training.
However, the primary objective is the TAD losses ($\mathcal{L}_{\text{DETR}}$), and therefore, the feedback losses do not overwhelm the objectives in the beginning of the learning.
Fig.~\ref{fig:losses} shows the training losses of the DETR and prediction-feedback.
As shown, the DETR loss decreases consistently and the feedback losses do not converge to 0.
This implies that our feedback objectives regularize the model to stay close to the predictions.
However, the attention maps do not reach the same as the target of the feedback, but they learn more expressive attention for the primary objective.

\vspace{3pt}
\noindent\textbf{THUMOS14 with VM2 Features.} 
Table.~\ref{tab:thumos14_vm2} shows the comparison results on THUMOS14 with VideoMAEv2-g features.
Interestingly, the performance gap between the standard methods and ours becomes much smaller compared to the results with the I3D features.
The results demonstrate that the better backbone features can compensate for the drawbacks of the DETR-based methods that cannot handle a long sequence for short actions.

\vspace{3pt}
\noindent\textbf{DETAD Analysis.} 
\begin{figure}[t]
\centering
\includegraphics[width=8.40cm]{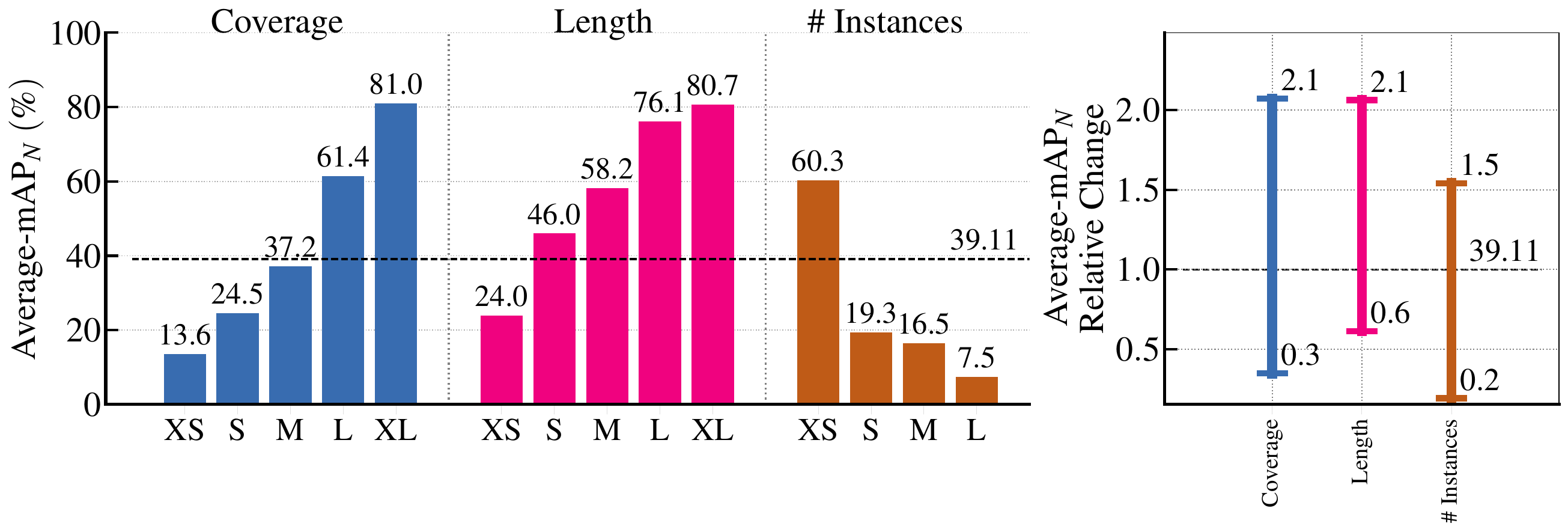}
\caption{\textbf{Feedback Losses.}
The figure shows the sensitivity analysis of DETAD on ActivityNet-v1.3.
}
\label{fig:detad_sensitivity}
\end{figure}
Fig.~\ref{fig:detad_sensitivity} shows the sensitivity analysis of DETAD~\cite{alwassel2018detad} on ActivityNet-v1.3.
As shown, Pred-DETR shows high performances over scales.